# Bees Local Phase Quantization Feature Selection for RGB-D Facial Expressions Recognition

Seyed Muhammad Hossein Mousavi

Atiye Ilanloo

**Abstract** Feature selection could be defined as an optimization problem and solved by bio-inspired algorithms. Bees Algorithm (BA) shows decent performance in feature selection optimization task. On the other hand, Local Phase Quantization (LPQ) is a frequency domain feature which has excellent performance on Depth images. Here, after extracting LPQ features out of RGB (colour) and Depth images from the Iranian Kinect Face Database (IKFDB), the Bees feature selection algorithm applies to select desired number of features for final classification tasks. IKFDB is recorded with Kinect sensor V.2 and contains colour and depth images for facial and facial micro-expressions recognition purposes. Here five facial expressions of Anger, Joy, Surprise, Disgust and Fear are used for final validation. The proposed Bees LPQ method is compared with Particle Swarm Optimization (PSO) LPQ, PCA LPQ, Lasso LPQ, and just LPQ features for classification tasks with Support Vector Machines (SVM), K-Nearest Neighbourhood (KNN), Shallow Neural Network and Ensemble Subspace KNN. Returned results, show a decent performance of the proposed algorithm (99 % accuracy) in comparison with others.

Seyed Muhammad Hossein Mousavi
Independent Researcher, Tehran, Iran, e-mail: mosavi.a.i.buali@gmail.com
Atiye Ilanloo
Faculty of Humanities- Psychology, Islamic Azad University-Rasht, Gilan, Iran, e-mail: elanlooatiye@gmail.com

# 1 Introduction

**1.1 Facial Expressions Recognition**

After face detection [1] and face recognition [2], analysing facial muscles' movement comes to mind. These movements are called Facial Expressions Recognition (FER) [2] and Facial Micro Expressions Recognition (FMER) [2] and have high importance in image processing and psychology. Recognising facial expressions is harder than face as they could have micro-movements and differs from one person to another. Also, each expression could be misunderstood in different races as skin wrinkles effects recognition direction. Facial expressions are made and calculated by Facial Action Coding Systems (FACS) [3] and they are categorized in seven primary facial expressions of Joy, Anger, Disgust, Sadness, Fear, Surprise and Neutral. For example, and for Joy expression, Action Units (AU) of + and 12 are involved. FMER is FER with more precise calculations as little changes appear on the face. Some of these expressions are illustrated in Fig 1 alongside with corresponding their action unts.

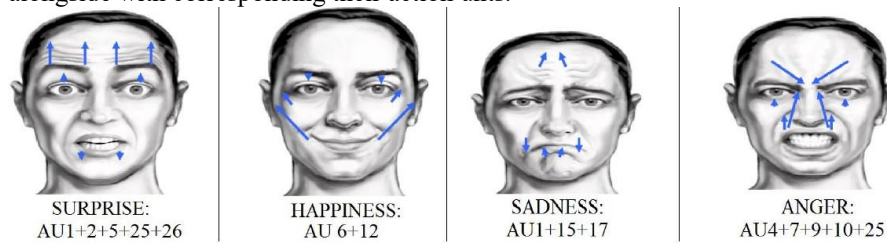

**Fig 1. Four main facial expressions**

**1.2 Bio-inspired Algorithms and Bees Algorithm**

Bio-inspired algorithms [4, 5] normally are mathematical models of animals and insects' social behaviour in a manner that leads problems to an optimal solution during iterations by specific number of populations in each generation. These algorithms could be employed in multiple optimization tasks, such as regression [6], clustering [7, 9], feature selection [8], Minimum Spanning Tree (MST), Hub Location Allocation (HLA) and more. One of these bio-inspired algorithms which has highly efficiency is called Bees Algorithm (BA) [10, 11, 18]. Bees algorithm has a lot of applications which Bees Algorithm, simply implements the social behaviour of honey bees to search (in neighbourhood manner) for food in flowers. Agents, Scouts and Forager bees are involved in global and local search to reach the best solution. Waggle dance in done by scouts which found the best sites. Those who landed on elite sites, recruit new members. List of best bees based on local and global goes to next generation and cycle stopes by termination criterions conditions.

### 1.3 Kinect Sensor and Depth Data

There lots of infrared sensors out there which Kinect sensor [12] is one most efficient and cheapest ones which uses for educational purposes. Here data recorded with Kinect Version 2 sensor is used in validation section. Kinect V.2 is capable of recording colour data with 1920*1080 resolutions and Depth data in 512*424 resolutions with 30 Frame Per Seconds (FPS). Colour data is based on Red, Green and Blue (RGB) channels and depth data is a grey like image which sometimes called 2.5-Dimensional (2.5-D) image. Each pixel of Depth image in Kinect sensor, represents the distance between Sensor and Subject in millimetres. The range is 0.8 meter to 5 meters and for instance 2000 value in pixels means object is in distance of 2 meters from the sensor. Depth images could be transformed to 3-D images and aid colour images to increase recognition accuracy. As depth images are generated from infrared particles, the Kinect sensor could operate in absolute darkness.

### 1.4 Feature Selection

In dealing with big data [13] or massive number of samples (specially image samples), and after the feature extraction [14] step, selecting the most impactful features out of data is essential. Feature selection [15] or dimensionality reduction helps to decrease computational time and volume plus getting rid of outliers in the dataset. Outliers lead classification [16] task into misclassification and removing them is vital. In this paper, Bees Algorithm is employed as a feature selection tool. Here, the Local Phase Quantization (LPQ) [17] feature is extracted which provides 256 features and Bees feature selection, select less than 128 features without classification accuracy drop. The paper is consisting of five main sections of introduction, prior related researches, proposed method, validation and results and conclusion.

## 2 Prior Related Researches

As this research is about feature selection, famous and benchmark feature selection or dimensionality reduction techniques belonging to traditional and bio-inspired category will be explained. These algorithms could be applied directly on extracted image or signal features or could be applied on unfolded version of images, signals or any type of numerical matrix to get rid of outliers and reduce computational time.

Principal Component Analysis (PCA) [19] generates new matrixes, named Principal Components (PA). Each PA is a linear mixed of the original matrixes. All the PAs are orthogonal to one another, so there is no extra information. By selecting best principles components, best features will be chosen.

On other hand however, the Lasso [20] is a regularization [20] method for estimating generalized linear models. Lasso includes a red line that limits the size of the approximated coefficients. So, it is like Ridge Regression [21]. Lasso is a shrinkage approximator: it makes coefficient approximated that are biased to be tiny. By shrinking features, best features remain.

The chi-square test [22] is a statistical one employed to contrast observed data and expected ones. The objective is to specify if a changes among observed data and expected data is because of chance, or if it is because of a relevance between the variables that are being studied. Having two variables, it is possible to get observed count and expected count. Chi-Square measures how desired count and observed count deviates each other. If the target variable is independent of the feature variable, it is possible to ignore that specific feature. If they are dependent, the feature is very significant.

Selecting best features based on the highest Data Envelopment Analysis (DEA) [23] for Data Management Units (DMUs) [23] is a management-related feature selection technique. In that, those features or DMU's with best efficiency will be selected and the rest will be considered as leftovers or outliers [24].

Among bio-inspired feature selection techniques, few has a proper performance which are includes PSO feature selection [25], Firefly feature selection [26] and Bees feature selection [27].

## 3 Proposed Method

The proposed method is consisted multiple parts of data acquisition, pre-processing, extracting LPQ features, selecting Bees features, labeling, classification and comparison. These operations are illustrated in Fig 2. For validation, the Iranian Kinect Face Data Base (IKFDB) [2] is employed which is consisted of 40 subjects. This data set contain facial images for face recognition, age estimation facial expressions and facial micro expressions recognition tasks. Also, data is recorded with Kinect sensor V.2 in color and depth formats. All main seven facial expressions along side with pitch, roll and yaw as video frames are available in this database. Here, five expressions of Joy, Anger, Fear, Disgust and Surprise in color and depth formats are used for the main experiment. After reading data images from input, pre-processing stage including, grey level conversion, intensity adjustment, histogram equalization and image resizing takes place. The next step is extracting LPQ feature from the frequency domain from all color and depth samples. For each image sample, 256 LPQ feature will be extracted. Fourth step is consisted of feature selection by Bees algorithm in desired number of features. Here the range of 2 to 255 feature could be selected. As classification task is supervised learning method, labeling is required for the fifth step. Classification algorithms are Support Vector Machines (SVM) [5], K-Nearest Neighborhood (KNN) [28], Shallow Neural Network [2] and Ensemble Subspace KNN [29]. Next, comparison by PCA, Lasso,

PSO feature selection and LPQ only using mentioned classification algorithms, takes place. Finally, related confusion matrixes and Receiver Operating Characteristic (ROC) [30] curves plots.

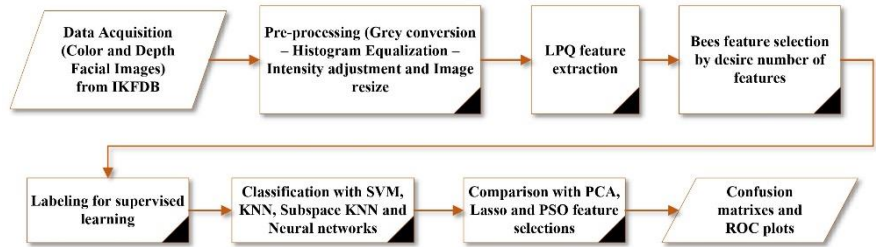

**Fig 2. Proposed method flowchart**

### 3.1 Local Phase Quantization

LPQ [17] is a frequency neighborhood-based feature based on Fourier transform [31]. It manipulated blurring effect in magnitude and phase channels. Phase channel is capable of deactivating low pass filters that exists in some images. LPQ features is perfect to be use on depth data in frequency domain. Figure 3 illustrates the LPQ algorithm process. Furthermore, this feature has perfect performance in dealing with low pass gaussian filters.

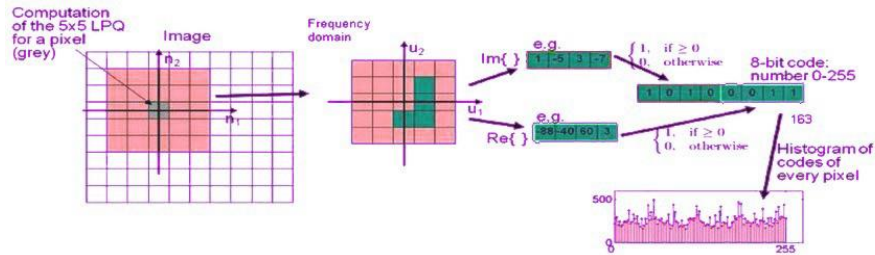

**Fig 3. LPQ algorithm workflow**

### 3.2 Bees Feature Selection

In feature selection, we are dealing with Number of Features of "NF", weight of feature or "w" and Mean Square Error (MSE) [32] which should be minimized to select thee feature. Also, if $x_i$ are values of NF then, $\hat{x_i}$ would be selected features out of NF. So, considering number of features entering the system, "y' would be the output and "t" would be the target. In order to calculate final error, $e_i$ need to be calculated which is $t_i$ - $y_i$ . So final error is min $MSE = \frac{1}{n}\sum_{i=1}^{n} e_i^2$ +w*NF. This goes for all features and finally those features with lowest MSE will be selected. In combination of Bees and feature selection, each feature vector is considered as a bee with different fitness function. Those bees which could fit into final iteration would be selected alongside with their related features with lowers error as it mentioned. Pseudo code of Bees feature selection is presented below.

*Start*
  *Load LPQ features*
    *Generating initial population (Features of F)*
    *Define FN (number of features) and w (wight of features)*
    *Evaluating the population based on fitness function*
    *Sorting*
      **While** *max iteration is not satisfied*
        *Select elite patches and non-elite best patches for local search*
        *Recruit forager bees for elite patches and non-elite best patches*
        *Evaluate the fitness value of each patch*
        *Sorting (select NF of F)*
        *Allocate the rest of the bees for global search*
        *Evaluate the fitness value of non-best patches*
        *Sorting (Select NF of F)*
      **End While**
    *Select best first NF's*
**End**

## 4 Validation and Results

As it mentioned earlier for validation, IKFDB [2] data is used which is consisted of color and depth frames of seven main facial expressions. Here 1000 color and depth samples of five expressions (each 200 samples) are used in this experiment.

Some samples of IKFDB in different expressions and in color-depth form is presented in Fig 4. Also, comparison would be between proposed Bees LPQ, PSO LPQ [25], PCA LPQ [19], Lasso LPQ [20] and solo LPQ using SVM [5], KNN [28], Shallow Neural Network [2] and Ensembles Subspace KNN [29] classification algorithms. Table 1 presents PSO and Bees parameters which is used in the experiment. Table 2 represents classification comparison results for different feature selection algorithms with half features (128), quarter of features (64) and octant of features (32). Obviously, all features (256) are considered for solo LPQ in this table.

Fig 5 represents, confusion matrixes and ROC curves for solo LPQ with all 256 features, Lasso algorithm with 64 features and Bees algorithm with 64 features for SVM classifier. Fig 6 illustrates the same but for KNN classifier. Fig 7, presents the performance of Bees algorithm for feature selection task over 100 iterations.

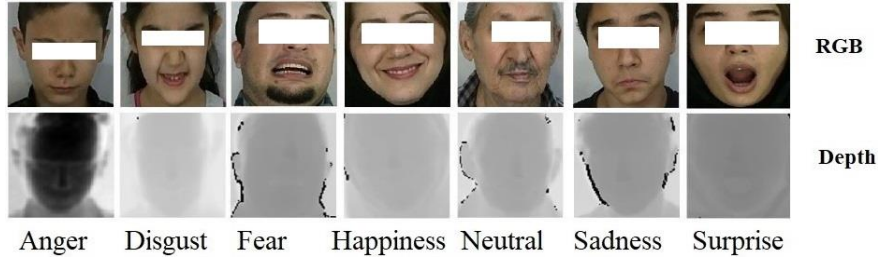

Fig 4. Samples of the IKFDB

Table 1. Bees and PSO Parameters

| Algorithm | Bees | PSO |
|---|---|---|
| Iteration | 100 | 100 |
| Population | 10 Bees | 20 Particles |
| Decision Variable | 20 | 20 |
| Decision Variable Size | [1, 20] | [1, 20] |
| Lower Bound (LB) | -10 | -10 |
| Upper Bound (UB) | 10 | 10 |
| Mutation Rate | 0.2 | 0.2 |
| Inertia Weight | - | 1 |
| Inertia Weight Damping Ratio | - | 0.99 |
| Personal Learning Coefficient | - | 1.5 |
| Global Learning Coefficient | - | 2 |
| Selected Sites (SS) | Bees * 0.5 | - |
| Select Elite Sites | SS * 0.4 | - |
| Recruited Bees for Selected Sites | Bees * 0.5 | - |
| Recruited Bees for Elite Sites | SS * 2 | - |
| Neighbourhood Radius | 0.1 * (UB – LB) | - |
| Neighbourhood Radius Damp Rate | 0.2   0.96 | - |

Looking at Table 2 results, best results belong to Bees features selection, PSO feature selection and PCA feature selection in different number of features. Additionally, Lasso feature selections place at the end of the ranking according to its weak performance. Also, solo LPQ with all 256 features places in the middle of ranking table.

PSO and Bees are competing with each other with different classifiers and best result ever belongs to Bees feature selection algorithm with 128 features and by subspace KNN classifier (99.8 % accuracy). However, the weakest performance goes to Lasso with 32 features and with SVM classifier (84.5 % accuracy). Here facial expressions are labels numerically as 1: Anger, 2: Joy, 3: Fear, 4: Disgust and 5: Surprise.

**Table 2. Comparison Classification Results with Different Number of Features (256, 128, 64 and 32)**

|  | LPQ | PCA | Lasso | PSO | Bees |
|---|---|---|---|---|---|
| **SVM** | 256 : 98.8 % | 128 : 98.4 % | 128 : 90.6 % | 128 : 99.1 % | 128 : 99.6 % |
|  |  | 64 : 98.0 % | 64 : 89.3 % | 64 : 98.6 % | 64 : 98.9 % |
|  |  | 32 : 90.8 % | 32 : 84.5 % | 32 : 97.4 % | 32 : 97.3 % |
| **KNN** | 256 : 98.0 % | 128 : 98.1 % | 128 : 93.0 % | 128 : 98.8 % | 128 : 99.2 % |
|  |  | 64 : 97.9 % | 64 : 93.9 % | 64 : 98.2 % | 64 : 98.1 % |
|  |  | 32 : 97.5 % | 32 : 92.8 % | 32 : 97.6 % | 32 : 97.7 % |
| **Shallow NN** | 256 : 97.6 % | 128 : 97.3 % | 128 : 96.7 % | 128 : 99.1 % | 128 : 99.4 % |
|  |  | 64 : 96.6 % | 64 : 90.7 % | 64 : 97.8 % | 64 : 98.7 % |
|  |  | 32 : 95.8 % | 32 : 91.2 % | 32 : 96.9 % | 32 : 97.9 % |
| **Ensemble Subspace KNN** | 256 : 98.3 % | 128 : 97.9 % | 128 : 95.6 % | 128 : 99.5 % | 128 : 99.8 % |
|  |  | 64 : 97.7 % | 64 : 93.1 % | 64 : 98.3 % | 64 : 98.9 % |
|  |  | 32 : 98.1 % | 32 : 89.1 % | 32 : 98.0 % | 32 : 98.5 % |

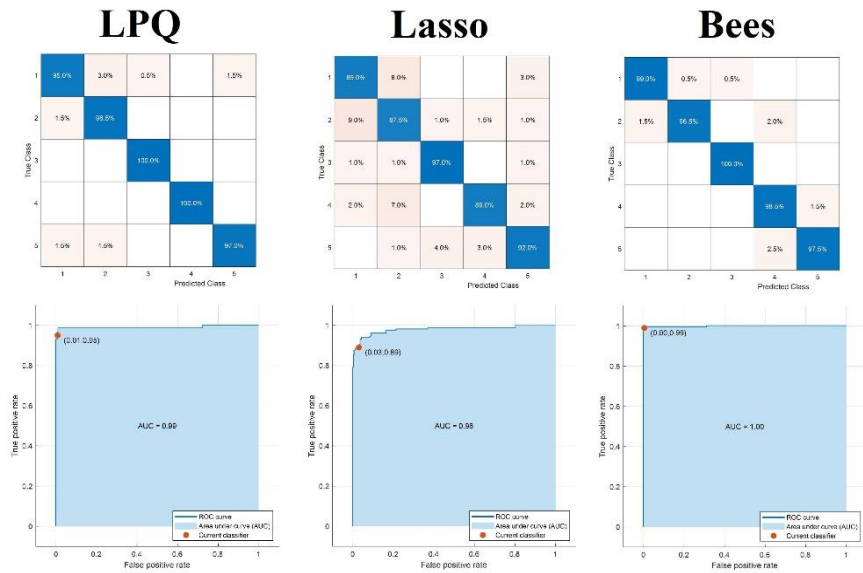

**Fig 5. Confusion matrxies and ROC curves for LPQ, Lasso and Bees algorithms with 256, 64 and 64 features with SVM classifier**

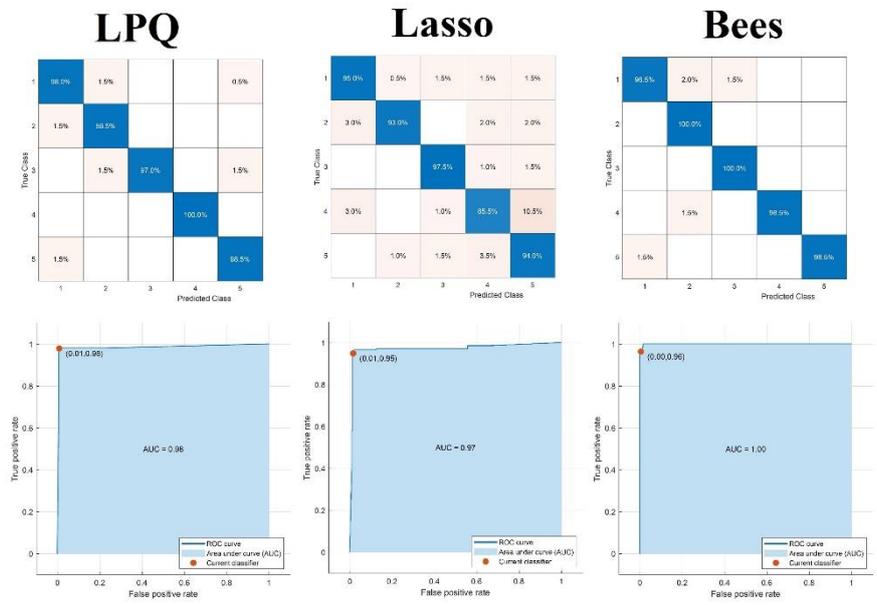

**Fig 6. Confusion matrxies and ROC curves for LPQ, Lasso and Bees algorithms with 256, 64 and 64 features with KNN classifier**

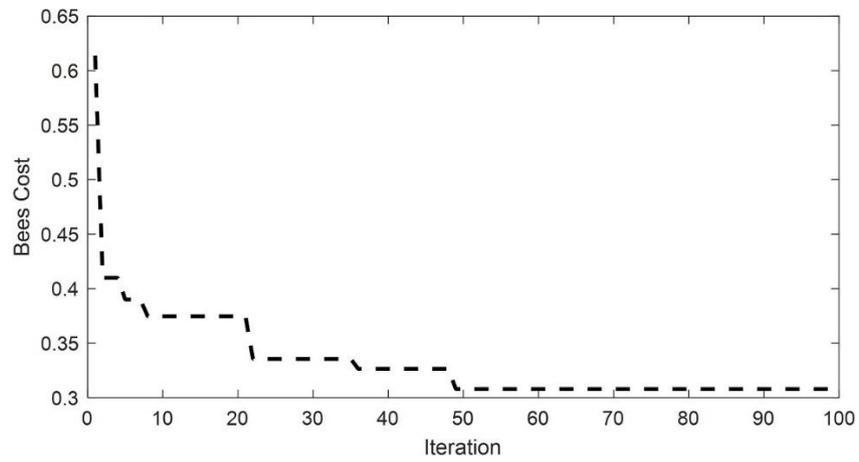

**Fig 7. Bees algorithm performance over 100 iteration for feature selection task**

## 4 Conclusion and suggestions

Employing bio-inspired algorithms for feature selection task could perform better than traditional feature selection algorithms in face analysis and especially facial expressions recognition task. Defining feature selection cost function for Bees algorithm returned successful results for all four classification algorithms. So, it can be concluded, Bees algorithm has great impact in feature selection optimization, after feature extraction by LPQ features. However, run time increases but, Bees algorithm with using just a quarter of features, could reach to a point which traditional algorithms are not capable of. Furthermore, using proposed method on more facial expressions and in other face analysis tasks such as face recognition is of future works. Also, it is suggested to use proposed method with more classification algorithms and comparing with more traditional and bio-inspired feature selection techniques.